\documentclass{article}
\pdfoutput=1

\usepackage[utf8]{inputenc} 
\usepackage[T1]{fontenc}    
\usepackage{url}            
\usepackage{booktabs}       
\usepackage{amsfonts}       
\usepackage{nicefrac}       
\usepackage{microtype}      

\usepackage{graphicx}
\usepackage{subcaption}
\usepackage{xcolor}
\usepackage{xr}

\usepackage{multicol}
\usepackage{lipsum}
\usepackage{mwe}
\usepackage{wrapfig}

\usepackage{authblk}

\title{Knowledge Graph Anchored Information Extraction for Domain-Specific Insights}

\author{Vivek Khetan}
\author{Annervaz K M}
\author{Erin Wetherley}
\author{Elena Eneva}
\author{Shubhashis Sengupta}
\author{Andrew E. Fano}
\affil{Accenture labs}
\affil{\textit {\{vivek.a.khetan,annervaz.k.m,erin.wetherley,elena.eneva, shubhashis.sengupta,andrew.e.fano\}@accenture.com}}
 \date{\vspace{-5ex}}

\begin{document}
\maketitle

\begin{abstract}

The growing quantity and complexity of data pose challenges for humans to consume information and respond in a timely manner. For businesses in domains with rapidly changing rules and regulations, failure to identify changes can be costly.

In contrast to expert analysis or the development of domain-specific ontology and taxonomies, we use a task-based approach for fulfilling specific information needs within a new domain. Specifically, we propose to extract task-based information from incoming instance data.

A pipeline constructed of state of the art NLP technologies, including a bi-LSTM-CRF model for entity extraction, attention-based deep Semantic Role Labeling, and an automated verb-based relationship extractor, is used to automatically extract an instance level semantic structure.

Each instance is then combined with a larger, domain-specific knowledge graph to produce new and timely insights. Preliminary results, validated manually, show the methodology to be effective for extracting specific information to complete end use-cases.

\end{abstract}
\paragraph{Introduction:}

\begin{figure}[ht!]
\centering
\includegraphics[width=\textwidth,
height=6cm]{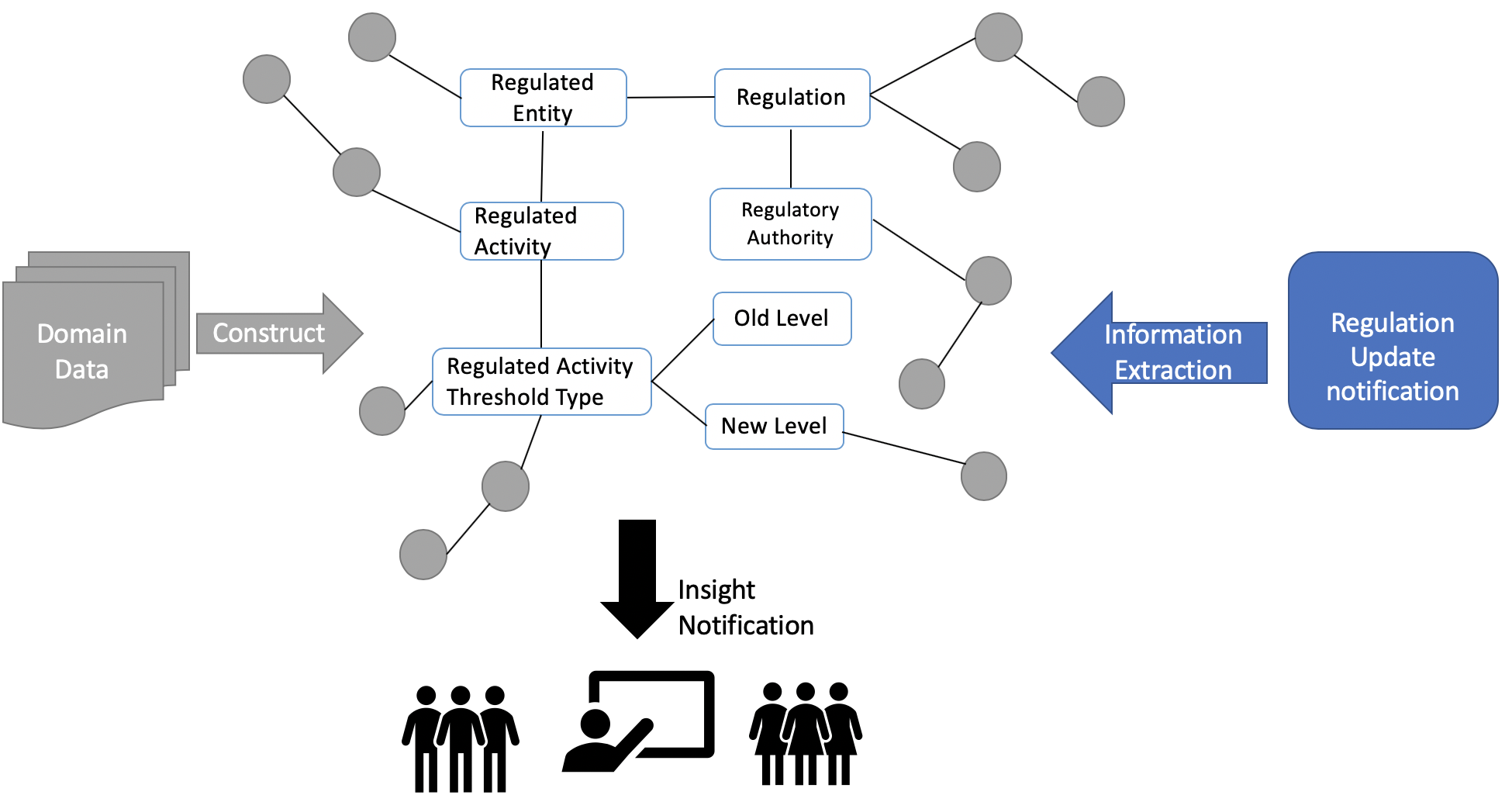}\quad
\caption{High-level solution schematic}
\label{fig:SolutionSchematic}
\end{figure}

The sheer growth in unstructured content is overwhelming the ability of businesses to respond effectively. For example, there are over 180,000 pages of regulation in the federal register and they are updated frequently. In the banking industry alone, the costs of staying compliant with (local to global) regulatory requirements are expected to exceed \$100 billion annually by 2020 \cite{regStats}. Staying on top of this depends on a combination of human and machine approaches. The goal of this study is to be able to extract and infer just enough to be able to focus human attention on the right content. For example, given news of a regulatory change, can we understand just enough to infer what businesses might be impacted and who needs to be notified.

This work at a basic level is an effort to ease up this information consumption need for specific tasks in a domain. 

We first construct a knowledge graph for the specific domain, in a semi-automated fashion, from an unstructured text corpus of the domain. With the help of a task-specific data model, we extract new information and provide it in an easy to consume manner on top of the constructed knowledge graph. The extracted information is also continuously updated in the knowledge graph, which helps to generate domain-related insights over a period of time. This is a second-level utility of the proposed approach.

Although we developed this approach for the regulatory domain, it is generic enough to be applied to various domain-specific tasks. Figure ~\ref{fig:SolutionSchematic} gives a high level view.



\paragraph{Study Data}

We assembled a dataset from three sources capturing the financial regulations and banking infrastructure of the United States. First, updates to U.S federal regulations were downloaded from the U.S. Federal Register (articles) \cite{fedReg}. Second, we identified existing financial regulations by downloading an XML file of Title XII of the U.S. Code of Federal Regulations \cite{title12}. Finally, we used a database (CSV) detailing all U.S. financial infrastructure (e.g., active banks, holding companies, federal regulators, etc.), downloaded from the National Information Center: Federal Financial Institutions Examination Council (NIC) \cite{nisData}.

\paragraph{Data-model development}

Here, we outline the development and application of a small, task-specific data-model within the context of monitoring financial regulatory change. We used the federal register articles to develop a domain-specific data-model that could identify and semantically relate a set of event-specific entities. The task targeted by the data-model developed for this study was to detect actors and events surrounding any change in a regulatory threshold, which we define as any increase or decrease in quantitative value that affects regulated banks. Data-model development was further informed by consultation with financial regulation subject matter experts. Task-specific entities in the data-model were manually identified and labelled within 131 articles. In total, we identified and extracted various instances of 7 entities from approximately 4500 sentences. The final number of labeled entity instances ranged from a minimum of 561 for regulated\_activity\_threshold, to a maximum of 6752 for regulatory\_authority.

\paragraph{Data Extraction} 
We trained several NLP models to automatically process a new article to extract entities and relationships and fill their relative slots in the data-model, including:
\begin{itemize}
    \item Semantic Role Labeling (SRL): We used the attention-based deep model of Tan et al. \cite{tan2018deep}. This RNN-based model first identifies and disambiguates a predicate, and then classifies all other tokens based on their roles. The output of an SRL process for a single document consists of many predicates, each paired with multiple actors.
    
    \item Custom Entity Extraction: A bi-directional LSTM model was trained on a combined dataset that included both our labelled data and CoNLL 2017 shared task data for named entities in BIO format \cite{conll-2017-conll-2017}. Combining datasets in this way has been shown to boost performance for datasets with limited annotation. 
    
    \item Relationship Extraction: We used the methodology of Q. Hao et al. \cite{hao2017verb} to automatically extract relationships based on the set of entities extracted from a given article. This method is a verb-based algorithm that can extract multiple relationships for a given pair of entities in an unstructured text document. Here, we extracted multiple relationships between our entities (if available) to capture different associations between them.

\end{itemize}

\begin{figure}[ht!]
\centering
  \includegraphics[width=\textwidth, height=8cm]{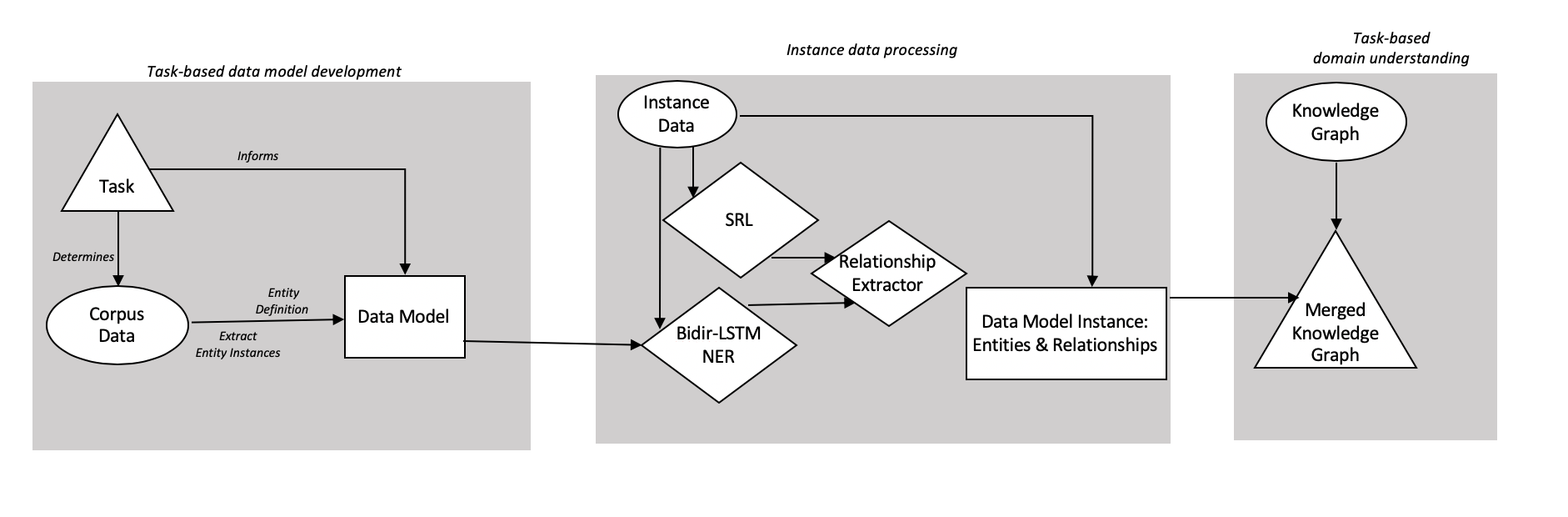}\quad
  \caption{Study flowchart}
  \label{fig:FlowChart}
\end{figure}

\paragraph{Pipeline for automated instance-level semantic structure extraction}
The above algorithms were integrated into a single pipeline, which was used to automatically extract news of a threshold change event from an instance-level data article as shown in Figure \ref{fig:FlowChart}. First, an article was simultaneously processed to extract custom entities using the custom entity extractor, and actor information using SRL. A final, filtered instance entity list consisted of all entities which were extracted by both methods. Entities that were only extracted by either SRL or the custom extractor were discarded. Next, every possible pair of the remaining entities were input into the relationship extractor algorithm and assessed against the article to extract all relevant relationships and form triples. These were used to fill the slots in the data-model, producing a data-model instance that semantically represented the article. 

\paragraph{Domain knowledge graph construction}
Each data-model instance represents a semantic knowledge graph with limited scope due to the small size of our data-model and lack of rich training data for our extractors. To overcome this limitation, we wrote heuristics based on OpenIE \footnote{\url{https://stanfordnlp.github.io/CoreNLP/openie.html}} and ClauseIE \cite{DelCorro:2013:CCO:2488388.2488420} extraction triplets, thus extracting more nodes and relationships from each article. These were combined with the data model instance. Finally, the domain-specific NIC dataset was incorporated into the knowledge graph, mapping relationships between regulatory agency, bank, bank branch, bank holding company, and federal insurance agency entities. Additional bank information was included as node properties, including addresses, bank assets, holdings, etc.


\paragraph{Generating notifications \& insights}

Notifications for a user, depicted in the knowledge graph, were generated in two ways. First, using subscription rules written manually based on properties of the extracted information and a user's provided information needs. Second, using the wordnet\footnote{\url{https://wordnet.princeton.edu}} based semantic similarity of the role description and the information metadata. A key challenge is to reason about the domain in a limited way with mixed quality information, and this is an ongoing effort. Based on their roles and responsibilities, users receive a notification with each relevant federal register update. The information is also used to update the knowledge graph. Over a period of time, this updated information helps to develop insights into events in the domain. This can also be consumed directly by the user by querying the knowledge graph.

\paragraph{Early results \& discussion}
Quantifying results in numerical terms is difficult for the problem we are addressing. In our use-case, low precision of the overall extraction is acceptable as users are able to easily discard any superfluous information that is returned by a high recall. A measure called summarization \cite{10.1007/978-3-642-28714-5_18}, defined as the ratio between the size of the input document and size of the output, has been designed to measure the reduction of human effort achieved through text processing. In our case, the summarization value is promising and we are currently evaluating it across our dataset.

\paragraph{Conclusion \& future work}

In this paper, we present a solution for semi-automatic consumption of continuous flows of information from large documents, with a focus on fulfilling certain domain-specific tasks and roles. This is based on the semi-automatic construction of a large domain knowledge graph from a domain text corpus, processing of new instance streams, and extracting task-based information using criteria set by users. 

This work is in its very early stages, with several opportunities for future direction. One area of key interest is expanding alert rules to include those automatically generated by machine learning of user browsing history. Another is overcoming the limited available knowledge graph training data using techniques such as REHession \cite{DBLP:journals/corr/LiuRZZGJ017}. Finally, knowledge graph completion- and reasoning-based methods \cite{DBLP:journals/corr/abs-1803-06581} can be applied to use-case specific reasoning tasks to generate hidden insights. We hope this initial work will inspire further study of the problem.

\bibliographystyle{unsrt}
\bibliography{RegTech.bib}

\begin{thebibliography}{10}

\bibitem{regStats}
Reg stats.
\newblock \url{https://regulatorystudies.columbian.gwu.edu/reg-stats}.
\newblock Accessed: 2019-09-05.

\bibitem{fedReg}
United states federal register api.
\newblock \url{https://www.federalregister.gov/developers/api/v1}.
\newblock Accessed: 2019-09-05.

\bibitem{title12}
Electronic code of federal regulations, title xii.
\newblock \url{https://www.govinfo.gov/bulkdata/ECFR/title-12}.
\newblock Accessed: 2019-09-05.

\bibitem{nisData}
National information center: Federal financial institutions examination
  council.
\newblock \url{https://www.ffiec.gov/npw/FinancialReport/DataDownload}.
\newblock Accessed: 2019-09-05.

\bibitem{tan2018deep}
Zhixing Tan, Mingxuan Wang, Jun Xie, Yidong Chen, and Xiaodong Shi.
\newblock Deep semantic role labeling with self-attention.
\newblock In {\em Thirty-Second AAAI Conference on Artificial Intelligence},
  2018.

\bibitem{conll-2017-conll-2017}
{\em Proceedings of the {C}o{NLL} 2017 Shared Task: Multilingual Parsing from
  Raw Text to Universal Dependencies}, Vancouver, Canada, August 2017.
  Association for Computational Linguistics.

\bibitem{hao2017verb}
Qi~Hao, Jeroen Keppens, and Odinaldo Rodrigues.
\newblock A verb-based algorithm for multiple-relation extraction from single
  sentences.
\newblock In {\em Proceedings of the International Conference on Information
  and Knowledge Engineering (IKE)}, pages 115--121. The Steering Committee of
  The World Congress in Computer Science, Computer~…, 2017.

\bibitem{DelCorro:2013:CCO:2488388.2488420}
Luciano Del~Corro and Rainer Gemulla.
\newblock Clausie: Clause-based open information extraction.
\newblock In {\em Proceedings of the 22Nd International Conference on World
  Wide Web}, WWW '13, pages 355--366, New York, NY, USA, 2013. ACM.

\bibitem{10.1007/978-3-642-28714-5_18}
Daniel Berry, Ricardo Gacitua, Pete Sawyer, and Sri~Fatimah Tjong.
\newblock The case for dumb requirements engineering tools.
\newblock In Bj{\"o}rn Regnell and Daniela Damian, editors, {\em Requirements
  Engineering: Foundation for Software Quality}, pages 211--217, Berlin,
  Heidelberg, 2012. Springer Berlin Heidelberg.

\bibitem{DBLP:journals/corr/LiuRZZGJ017}
Liyuan Liu, Xiang Ren, Qi~Zhu, Shi Zhi, Huan Gui, Heng Ji, and Jiawei Han.
\newblock Heterogeneous supervision for relation extraction: {A} representation
  learning approach.
\newblock {\em CoRR}, abs/1707.00166, 2017.

\bibitem{DBLP:journals/corr/abs-1803-06581}
Wenhu Chen, Wenhan Xiong, Xifeng Yan, and William~Yang Wang.
\newblock Variational knowledge graph reasoning.
\newblock {\em CoRR}, abs/1803.06581, 2018.

\end{thebibliography}
\medskip

\small


\end{document}